\documentclass[runningheads]{llncs}
\usepackage[utf8]{inputenc}
\usepackage{graphicx}
\usepackage{textcomp}
\usepackage{xcolor}
\usepackage{booktabs}
\usepackage{amsmath}
\usepackage{amsfonts}
\usepackage{algorithm}
\usepackage{algpseudocode}
\usepackage{csquotes}
\usepackage{enumerate}
\usepackage{multirow}
\usepackage{multicol}
\usepackage{tabularx}
\usepackage{array}
\usepackage{xcolor}
\usepackage{subcaption}
\usepackage{flushend}
\usepackage{comment}
\usepackage{mathrsfs}
\usepackage{gensymb}
\usepackage{tikz}
\usepackage{cite}
\usetikzlibrary{trees}

\usepackage{hyperref} 
\hypersetup{
    colorlinks=true,
    linkcolor=blue,
    filecolor=blue,      
    urlcolor=blue,
    citecolor=blue,
    }

\def\eg{\emph{e.g.}}
\def\ie{\emph{i.e.}}
\def\cf{\emph{cf.}}
\def\etal{\textit{et al.}}
\def \calM{{\mathcal{M}}}

\begin{document}
\sloppy

\title{(Local) Differential Privacy has NO Disparate Impact on Fairness\thanks{Version of Record (DBSec'23): \url{https://doi.org/10.1007/978-3-031-37586-6_1}.}}
\titlerunning{(Local) Differential Privacy has NO Disparate Impact on Fairness}
%
\author{
Héber H. Arcolezi \and
Karima Makhlouf \and
Catuscia Palamidessi
}

\authorrunning{H.H. Arcolezi, K. Makhlouf, \& C. Palamidessi
}

\institute{Inria and École Polytechnique (IPP), Palaiseau, France\\
\email{\{heber.hwang-arcolezi,karima.makhlouf,catuscia\}@lix.polytechnique.fr}
}

\maketitle              

\begin{abstract}
In recent years, Local Differential Privacy (LDP), a robust privacy-preserving methodology, has gained widespread adoption in real-world applications. With LDP, users can perturb their data on their devices before sending it out for analysis. 
However, as the collection of multiple sensitive information becomes more prevalent across various industries, collecting a single sensitive attribute under LDP may not be sufficient.
Correlated attributes in the data may still lead to inferences about the sensitive attribute.
This paper empirically studies the impact of collecting multiple sensitive attributes under LDP on fairness.
We propose a novel privacy budget allocation scheme that considers the varying domain size of sensitive attributes. 
This generally led to a better privacy-utility-fairness trade-off in our experiments than the state-of-art solution.
Our results show that LDP leads to slightly improved fairness in learning problems without significantly affecting the performance of the models. We conduct extensive experiments evaluating three benchmark datasets using several group fairness metrics and seven state-of-the-art LDP protocols.
Overall, this study challenges the common belief that differential privacy necessarily leads to worsened fairness in machine learning.

\end{abstract}

\section{Introduction}

The advent of the Big Data era has brought many benefits but has also raised significant concerns about privacy and algorithm bias in Machine Learning (ML).  
On the one hand, with massive amounts of data generated and collected by various entities, protecting individuals' personal information has become increasingly challenging.
In this context, research communities have proposed different methods to preserve privacy, with $\epsilon$-differential privacy ($\epsilon$-DP)~\cite{Dwork2006} standing out as a formal definition that allows quantifying the privacy-utility trade-off with the parameter $\epsilon$ (the smaller, the more private).
At the same time, there have been many efforts to develop methods and metrics to evaluate and promote fairness in ML due to unequal treatments of individuals or groups based on factors such as race, gender, or socio-economic status~\cite{alves2022survey,Makhlouf2021b,Makhlouf2021,Mehrabi2021}. 

This means that privacy and fairness are essential for ML to apply in practice successfully.
In real-life scenarios, it is not common anymore for entities to have access to \textit{sensitive} (or \textit{protected}\footnote{
Throughout this paper, we use the term \textit{sensitive} attribute from a privacy perspective and the term \textit{protected} attribute from a fairness perspective. Note that we always consider \textit{protected} attributes as \textit{sensitive} attributes.}) attributes like race due to legal restrictions and regulations\footnote{For example, the General Data Protection Regulation (GDPR)~\cite{GDPR}.} governing their collection.
Therefore, it can be difficult for these entities to quantify/assess the fairness of the models they deploy since they cannot access the protected attributes used for the fairness assessment.
One way to address this problem~\cite{Mozannar2020}, ignoring legal feasibility, is to enable users to share their sensitive attributes using protocols satisfying Local Differential Privacy (LDP)~\cite{first_ldp}, and learn a non-discriminatory predictor.

However, while collecting the sensitive attribute in a privacy-preserving manner may seem sufficient, it is worth noting that proxy variables can exist~\cite{Kallus2022} and can still lead to inferences about the sensitive attribute (\eg, by exploiting correlations). 
It is also important to acknowledge that proxy variables may be considered as personal information under the GDPR, requiring the same level of privacy protection. 
Thus, as collecting multiple sensitive information (\ie, \textit{multidimensional data}) becomes increasingly prevalent in various industries, protecting this information is a legal obligation and an ethical responsibility.

Therefore, this paper contributes to an in-depth empirical analysis of how pre-processing multidimensional data with $\epsilon$-LDP affects the fairness and utility in ML binary classification tasks.
We evaluated several group fairness metrics~\cite{Makhlouf2021,alves2022survey}, including disparate impact~\cite{barocas2016big}, equal opportunity~\cite{hardt2016equality}, and overall accuracy~\cite{berk2018fairness}, on benchmark datasets, namely, Adult~\cite{ding2021retiring}, ACSCoverage~\cite{ding2021retiring}, and LSAC~\cite{wightman1998lsac}.
To broaden the scope of our study, we have experimentally assessed seven state-of-the-art LDP protocols, namely, Generalized Randomized Response (GRR)~\cite{kairouz2016discrete}, Binary Local Hashing (BLH)~\cite{Bassily2015}, Optimal Local Hashing (OLH)~\cite{tianhao2017}, RAPPOR~\cite{rappor}, Optimal Unary Encoding (OUE)~\cite{tianhao2017}, Subset Selection (SS)~\cite{wang2016mutual,Min2018}, and Thresholding with Histogram Encoding (THE)~\cite{tianhao2017}. 

Moreover, since proxy variables can still introduce unintended biases and thus lead to unfair decisions\cite{Kallus2022}, we consider the setting in which each proxy (sensitive attribute) is collected independently under LDP guarantees. 
In other words, applying this independent setting automatically removes the correlation between the proxy attributes. 
To this end, the privacy budget $\epsilon$ should be divided among all sensitive attributes to ensure $\epsilon$-LDP under sequential composition~\cite{dwork2014algorithmic}.
Let $d_{s}$ be the total number of sensitive attributes, the LDP literature for multidimensional data~\cite{wang2019,Arcolezi2021_rs_fd} considers a \textbf{uniform} solution that collects each sensitive attribute under $\frac{\epsilon}{d_{s}}$-LDP.
In this paper, we propose a new \textbf{k-based} solution that considers the varying domain size $k$ of different sensitive attributes.
More precisely, for the $j$-th sensitive attribute, we allocate $\epsilon_j = \frac{\epsilon \cdot k_j}{\sum_{i=1}^{d_{s}} k_i}$.

Overall, our study challenges the common belief that using DP necessarily leads to worsened fairness in ML~\cite{Bagdasaryan2019,Ganev2022}. 
Our findings show that training a classifier on LDP-based multidimensional data slightly improved fairness results without significantly affecting classifier performance.
We hope this work can aid practitioners in collecting multidimensional user data in a privacy-preserving manner by providing insights into which LDP protocol and privacy budget-splitting solutions are best suited to their needs.

In summary, the three main contributions of this paper are: 
\begin{itemize}
    \item We empirically analyze the impact of pre-processing multidimensional data with $\epsilon$-LDP on fairness and utility;

    \item We compare the impact of seven state-of-the-art LDP protocols under a homogeneous encoding when training ML binary classifiers (see Fig.~\ref{fig:overview_encoding}) on fairness and utility;

    \item We propose a new privacy budget splitting solution named k-based, which generally led to a better privacy-utility-fairness trade-off in our experiments. 
\end{itemize}

All our codes are available in a \textbf{GitHub repository}~\cite{artifact_fairness_ldp}.

\textbf{Outline.} The rest of this paper is organized as follows.
Section~\ref{sec:rel_work} discusses related work.
In Section~\ref{sec:background}, we present the notation, fairness, and LDP protocols used.
Next, Section~\ref{sec:problem_setting_methodology} states the problem addressed in this paper and the proposed k-based solution.
Section~\ref{sec:experimental_evaluation} details the experimental setting and main results.
Finally, we conclude this work indicating future perspectives in Section~\ref{sec:conclusion}.

\section{Related Work} \label{sec:rel_work}

The recent survey work by Fioretto \etal~\cite{Fioretto2022} discusses two views about the relationship between central DP and fairness in learning and decision tasks.
The first view considers DP and fairness in an aligned space (\eg,~\cite{Dwork2012}), which mainly corresponds to individual fairness metrics.
The other view regards DP and fairness as ``enemies" (\eg,~\cite{Bagdasaryan2019,Ganev2022,Pujol2020}), which mainly corresponds to group fairness notions.
For instance, Pujol \etal~\cite{Pujol2020} investigated disparities in decision tasks using $\epsilon$-DP data.
Regarding learning tasks, Bagdasaryan, Poursaeed, \& Shmatikov~\cite{Bagdasaryan2019} studied the impact of training $\epsilon$-DP deep learning (\textit{a.k.a. gradient perturbation}) models on unprivileged groups.
By keeping the same hyperparameters as the non-private baseline model, the authors noticed that the accuracy for the unprivileged group dropped more than for the privileged one.
Similarly, Ganev et al.~\cite{Ganev2022} have also noticed disparities for the unprivileged group when generating $\epsilon$-DP synthetic data for training ML models by also keeping default hyperparameters of the differentially private generative models.
In this paper, we aim to explore to what extent training an ML classifier on $\epsilon$-LDP multidimensional data (\textit{a.k.a. input perturbation}) while fixing the same set of hyperparameters negatively impacts the unprivileged group is valid.

Regarding the local DP setting, the work of Mozannar, Ohannessian, \& Srebro~\cite{Mozannar2020} was the first one to propose a fair classifier when sanitizing only the protected attribute with $\epsilon$-LDP in both training and testing sets.
More recently, the work of Chen \etal~\cite{chen2022fair} considers a ``semi-private" setting in which a small portion of users share their protected attribute with no sanitization and all other users apply an $\epsilon$-LDP protocol.
While the two aforementioned research works~\cite{Mozannar2020,chen2022fair} answer interesting questions by collecting a single sensitive attribute using only the GRR~\cite{kairouz2016discrete} protocol, we consider in this work multiple sensitive attributes, which reflects real-world data collections, seven $\epsilon$-LDP protocols, and several fairness and utility metrics. 
In addition, we also propose a new privacy budget splitting solution named k-based, which generally leads to better fairness and performance in ML binary classification tasks. 

\section{Preliminaries and Background} \label{sec:background}

This section briefly reviews the group fairness metrics, LDP, and LDP protocols.
The notation used throughout this paper is summarized in Table~\ref{tab:notation}.

\begin{table}[ht]
    \centering
    \begin{tabular}{c c}
    \hline
         Symbol & Description \\\hline
         $n$ & Number of users \\
         $[n]$ & Set of integers, $\{1,2,\ldots,n\}$\\
         $\textbf{x}_i$ & $i$-th coordinate of vector $\textbf{x}$\\
         $z = \calM(v)$ & Protocol $\calM$ perturbs $v$ into $z$ under $\epsilon$-LDP\\
         $X$ & Set of ``non-sensitive" attributes\\         
         $A_{s}$ & Set of sensitive attributes (\textbf{privacy viewpoint}) \\
         $A_{p}$ & Protected attribute (\textbf{fairness viewpoint}), $A_{p} \in A_{s}$ \\
         $Z_{s}$ & Set of locally differentially private sensitive attributes, $Z_{s} = \calM(A_{s})$ \\
         $k_j$ & Domain size of the $j$-th attribute \\
         $d_s$ & Number of sensitive attributes, $d_s = |A_{s}| $ \\
         $Y$ & Set of target values, $Y=\{0, 1\}$ \\
         $D$ & Original dataset, $D=(X,A_{s},Y)$ \\
         $D_{z}$ & Dataset with sanitized sensitive attributes, $D_{z}=(X,Z_{s},Y)$ \\
         \hline
    \end{tabular}
    \caption{Notations}
    \label{tab:notation}
\end{table}

Note that in this work, we always consider a single protected attribute and assess fairness w.r.t. that attribute. For LDP, we consider a set of sensitive attributes instead. 
Moreover, the protected attribute is always considered sensitive, but the opposite is untrue. 

\subsection{Group Fairness Metrics} \label{sub:fairness_metrics}
In this paper, we focus on group fairness metrics, which assess the fairness of ML models for different demographic groups that differ by the protected attribute (\eg, race, gender, age, \dots). 
Let $A_{p}$ be the protected attribute, $\hat{Y}$ be a predictor of a binary target $Y \in \{0, 1\}$. The metrics we use to evaluate fairness are:

\begin{itemize}    
    \item \textbf{Disparate Impact (DI)}~\cite{barocas2016big}\textbf{.} DI is defined as the ratio of the proportion of positive predictions ($\hat{Y}=1$) for the \emph{unprivileged} group ($A_{p} = 0$) over the ratio of the proportion of positive predictions for the \emph{privileged} group ($A_{p} = 1$). The formula for DI is:
  
    \begin{equation} \label{eq:di}
        \textrm{DI} = \frac{\Pr[\hat{Y}=1 | A_{p} = 0]}{\Pr[\hat{Y}=1 | A_{p} = 1]} \textrm{.}
    \end{equation}
Note that a perfect DI value is equal to $1$.

    \item \textbf{Statistical Parity Difference (SPD)~\cite{agarwal2022fairness}}\textbf{.} Instead of the ratio, SDP computes the difference in the proportion of positive predictions for \emph{unprivileged} and \emph{privileged} groups and is defined as:
    \begin{equation} \label{eq:spd}
        \textrm{SPD} =  \Pr[\hat{Y}=1 | A_{p} = 1] - \Pr[\hat{Y}=1 | A_{p} =0]  \textrm{.}
    \end{equation}  
A perfect SPD value is equal to $0$.

    \item \textbf{Equal Opportunity Difference (EOD)~\cite{hardt2016equality}}\textbf{.} EOD measures the difference between the true positive rates (\ie, recall) of the \emph{unprivileged} group and the \emph{privileged} groups. Formally, EOD is defined as: 
    \begin{equation} \label{eq:eod}
        {\textrm{EOD} =  \Pr[\hat{Y}=1 | Y=1, A_{p} = 1] - \Pr[\hat{Y}=1 | Y=1, A_{p} =0]  \textrm{.}}
    \end{equation}  
A perfect EOD value is equal to $0$.

    \item \textbf{Overall Accuracy Difference (OAD)~\cite{berk2018fairness}}\textbf{.} OAD measures the difference between the overall accuracy rates between the \emph{privileged} group and the \emph{unprivileged} group. 
    Formally, OAD is represented as:
    \begin{equation} \label{eq:oad}
        \textrm{OAD} =  \Pr[\hat{Y}=Y | A_{p} = 1] - \Pr[\hat{Y}=Y | A_{p} =0]  \textrm{.}
    \end{equation}  
    A perfect OAD value is equal to $0$.
\end{itemize}

\subsection{Local Differential Privacy} 
\label{sub:ldp}
In this article, we use LDP~\cite{first_ldp} as the privacy model, which is formalized as:

\begin{definition}[$\epsilon$-Local Differential Privacy]\label{def:ldp} A randomized algorithm ${\calM}$ satisfies $\epsilon$-local-differential-privacy ($\epsilon$-LDP), where $\epsilon>0$, if for any pair of input values $v_1, v_2 \in Domain(\calM)$ and any possible output $z$ of ${\calM}$:

\begin{equation*} \label{eq:ldp}
    \Pr[{\calM}(v_1) = z] \leq e^\epsilon \cdot \Pr[{\calM}(v_2) = z]  \textrm{.}
\end{equation*}
\end{definition}

\begin{proposition}[Post-Processing~\cite{dwork2014algorithmic}]
\label{prop:post-processing}
If $\calM$ is $\epsilon$-LDP, then for any function $f$, the composition of $\calM$ and $f$, \ie, $f (\calM)$ satisfies $\epsilon$-LDP.
\end{proposition}

\begin{proposition}[Sequential Composition~\cite{dwork2014algorithmic}]
\label{prop:seq_comp}
Let $\calM_1$ be an $\epsilon_1$-LDP protocol and $\calM_2$ be an $\epsilon_2$-LDP protocol. 
Then, the protocol $\calM_{1,2}(v)=\left( \calM_1(v), \calM_2(v)\right)$ is $(\epsilon_1+\epsilon_2)$-LDP.
\end{proposition}

\subsection{LDP Protocols} 
\label{sub:ldp_protocols}

Let $A_{s}=\{v_1,\ldots,v_k\}$ be a sensitive attribute with a discrete domain of size $k=|A_{s}|$, in this subsection, we briefly review seven state-of-the-art LDP protocols. 
 
\subsubsection{Generalized Randomized Response (GRR)} 
\label{sub:GRR}

GRR~\cite{kairouz2016discrete} uses no particular encoding. 
Given a value $v \in A_{s}$, $GRR(v)$ outputs the true value $v$ with probability $p$, and any other value $v' \in A_{s} \setminus \{v\}$, otherwise. 
More formally:

\begin{equation*}
    \forall{z \in A_{s}}  : \quad \Pr[z=a] = \begin{cases} p=\frac{e^{\epsilon}}{e^{\epsilon}+k-1} \textrm{ if } z = a\\ q=\frac{1}{e^{\epsilon}+k-1} \textrm{ otherwise} \textrm{,} \end{cases}
\end{equation*}

\noindent in which $z$ is the perturbed value sent to the server. 

\subsubsection{Binary Local Hashing (BLH)} \label{sub:blh}

Local Hashing (LH) protocols~\cite{Bassily2015,tianhao2017} can handle a large domain size $k$ by first using hash functions to map an input value to a smaller domain of size $g$ (typically $2\leq g \ll k)$, and then applying GRR to the hashed value. 
Let $\mathscr{H}$ be a universal hash function family such that each hash function $H \in \mathscr{H}$ hashes a value in $A_{s}$ into $[g]$, \ie, $H : A_{s} \rightarrow [g]$. 
With BLH, $[g]= \{0, 1\}$, each user selects at random one hash function $H$, calculates $b= H(v)$, and perturbs $b$ to $z$ as:

\begin{equation*}
    \Pr[ z=1] = \begin{cases} p=\frac{e^{\epsilon}}{e^{\epsilon}+1} \textrm{ if } b = 1\\ q=\frac{1}{e^{\epsilon}+1} \textrm{ if } b=0 \textrm{.} \end{cases}
\end{equation*}

The user sends the tuple $\langle H, z \rangle$, \ie, the hash function and the perturbed value.
Thus, for each user, the server can calculate $S\left(\langle H, z \rangle\right) = \{v | H(v) = z \}$. 

\subsubsection{Optimal LH (OLH)} \label{sub:olh}

To improve the utility of LH protocols, Wang \etal~\cite{tianhao2017} proposed OLH in which the output space of the hash functions in family $\mathscr{H}$ is no longer binary as in BLH.
Thus, with OLH, $g=\lfloor e^{\epsilon} + 1 \rceil$, each user selects at random one hash function $H$, calculates $b= H(v)$, and perturbs $b$ to $z$ as:

\begin{equation*}
    \forall{i \in [g]}  : \quad \Pr[z=i] = \begin{cases} p=\frac{e^{\epsilon}}{e^{\epsilon}+g-1} \textrm{ if } b = i\\ q=\frac{1}{e^{\epsilon}+g-1} \textrm{ if } b \neq i \textrm{.} \end{cases}
\end{equation*}

Similar to BLH, the user sends the tuple $\langle H, z \rangle$ and, for each user, the server can calculate $S\left(\langle H, z \rangle\right) = \{v | H(v) = z \}$.

\subsubsection{RAPPOR} \label{sub:rappor}

The RAPPOR~\cite{rappor} protocol uses One-Hot Encoding (OHE) to interpret the user's input $v \in A_{s}$ as a one-hot $k$-dimensional vector.
More precisely, $\textbf{v}=OHE(v)$ is a binary vector with only the bit at position $v$ set to $1$ and the other bits set to $0$. 
Then, RAPPOR randomizes the bits from $\textbf{v}$ independently to generate $\textbf{z}$ as follows:

\begin{equation*} 
    \forall{i \in [k]} : \quad \Pr[\textbf{z}_i=1] =\begin{cases} p=\frac{e^{\epsilon/2}}{e^{\epsilon/2}+1} \textrm{ if } \textbf{v}_i=1, \\ q=\frac{1}{e^{\epsilon/2}+1} \textrm{ if } \textbf{v}_i=0 \textrm{,}\end{cases}
\end{equation*}

\noindent where $p+q=1$ (\ie, symmetric).
Afterwards, the user sends $\textbf{z}$ to the server.

\subsubsection{Optimal Unary Encoding (OUE)} \label{sub:oue}

To minimize the variance of RAPPOR, Wang \etal~\cite{tianhao2017} proposed OUE, which perturbs the $0$ and $1$ bits asymmetrically, \ie, $p+q\neq1$.
Thus, OUE generates $\textbf{z}$ by perturbing $\textbf{v}$ as follows:

\begin{equation*}  
     \forall{i \in [k]} : \quad \Pr[\textbf{z}_i=1] =\begin{cases} p=\frac{1}{2} \hspace{0.45cm} \textrm{ if } \textbf{v}_i=1, \\ q=\frac{1}{e^{\epsilon}+1} \textrm{ if } \textbf{v}_i=0 \textrm{.}\end{cases}
\end{equation*}

Afterwards, the user sends $\textbf{z}$ to the server.

\subsubsection{Subset Selection (SS)} \label{sub:SS}

The SS~\cite{wang2016mutual,Min2018} protocol randomly selects $1 \leq \omega \leq k$ items within the input domain to report a subset of values $\Omega \subseteq A_{s}$. 
The user's true value $v$ has higher probability of being included in the subset $\Omega$, compared to the other values in $A_{s} \setminus \{v\}$. 
The optimal subset size that minimizes the variance is $\omega= \lfloor \frac{k}{e^{\epsilon}+1} \rceil$. 
Given a value $v \in A_{s}$, $SS(v)$ starts by initializing an empty subset $\Omega$. 
Afterwards, the true value $v$ is added to $\Omega$ with probability $p=\frac{\omega e^{\epsilon}}{\omega e^{\epsilon} + k - \omega}$. 
Finally, it adds values to $\Omega$ as follows:

\begin{itemize}
    \item If $v \in \Omega$, then $\omega - 1$ values are sampled from $A_{s} \setminus \{v\}$ uniformly at random (without replacement) and are added to $\Omega$;
    
    \item If $v \notin \Omega$, then $\omega$ values are sampled from $A_{s} \setminus \{v\}$ uniformly at random (without replacement) and are added to $\Omega$.
\end{itemize}

Afterwards, the user sends the subset $\Omega$ to the server.

\subsubsection{Thresholding with Histogram Encoding (THE)} \label{sub:the}

Histogram Encoding (HE)~\cite{tianhao2017} encodes the user value as a one-hot $k$-dimensional histogram, \ie, $\textbf{v}=[0.0, 0.0, \ldots, 1.0, 0.0, \ldots, 0.0]$ in which only the $v$-th component is $1.0$.
$HE(\textbf{v})$ perturbs each bit of $\textbf{v}$ independently using the Laplace mechanism~\cite{Dwork2006}. 
Two different input values $v_1,v_2 \in A_{s}$ will result in two vectors with L1 distance of $\Delta=2$. 
Thus, HE will output $\textbf{z}$ such that $\textbf{z}_i = \textbf{v}_i + \textrm{Lap}\left( \frac{2}{\epsilon} \right)$.
To improve the utility of HE, Wang \etal~\cite{tianhao2017} proposed THE such that the user reports (or the server computes): $S(\textbf{z}) = \{ v \hspace{0.1cm} | \hspace{0.1cm} \textbf{z}_v \hspace{0.1cm} >  \hspace{0.1cm}\theta\}$, in which $\theta$ is the threshold with optimal value in $(0.5, 1)$.
In this work, we use \texttt{scipy.minimize\_scalar} to optimize $\theta$ for a fixed $\epsilon$ as: $\underset{\theta \in (0.5, 1)}{\min} \quad \frac{2 e^{\epsilon \theta / 2} - 1}{(1 + e^{\epsilon(\theta - 1/2)} - 2e^{\epsilon \theta / 2})^2}$.

\section{Problem Setting and Methodology} \label{sec:problem_setting_methodology}

We consider the scenario in which the server collects a set of multiple sensitive attributes $A_{s}$ under $\epsilon$-LDP guarantees from $n$ distributed users $U=\{u_1,\ldots,u_n\}$.
Furthermore, in addition to the LDP-based multidimensional data, we assume that the users will also provide non-sanitized data $X$, which we consider as ``non-sensitive" attributes.
The server aims to use both sanitized $Z_{s}=\calM(A_{s})$ and non-sanitized data $X$ to train an ML classifier with a binary target variable $Y=\{0,1\}$.
Notice, however, that we will be training an ML classifier on $D_{z} = (X,Z_{s},Y)$ but testing on $D = (X,A_{s},Y)$ as the main goal is to \textit{protect the privacy of the data used to train the ML model} (\eg, to avoid membership inference attacks~\cite{Hu2022}, reconstruction attacks~\cite{salem2020updates}, and other privacy threats~\cite{Liu2021}).
In other words, instead of considering a system for on-the-fly LDP sanitization of test data, as in~\cite{Mozannar2020}, we only sanitize the training set. 

With these elements in mind, our primary goal is to study the impact of training an ML classifier on $D_{z} = (X,Z_{s},Y)$ compared to $D = (X,A_{s},Y)$ on fairness and utility, using different LDP protocols and privacy budget splitting solutions.
More precisely, we consider the setting where each sensitive attribute in $A_{s}$ is collected independently under LDP guarantees.
In this case, to satisfy $\epsilon$-LDP following Proposition~\ref{prop:seq_comp}, the privacy budget $\epsilon$ must be split among the total number of sensitive attributes $d_{s}=|A_{s}|$.
To this end, the state-of-the-art~\cite{Arcolezi2021_rs_fd,wang2019} solution, named \textbf{uniform}, propose to split the privacy budget $\epsilon$ uniformly among all attributes, \ie, allocating $\frac{\epsilon}{d_{s}}$ for each attribute.
However, as different sensitive attributes have different domain sizes $k_j$, for $j \in [d_{s}]$, we propose a new solution named \textbf{k-based} that splits the privacy budget $\epsilon$ proportionally to the domain size of the attribute.
That is, for the $j$-th attribute, we will allocate $\epsilon_j = \frac{\epsilon \cdot k_j}{\sum_{i=1}^{d_{s}} k_i}$.

\begin{figure}[!ht]
    \centering
    \includegraphics[width=1\linewidth]{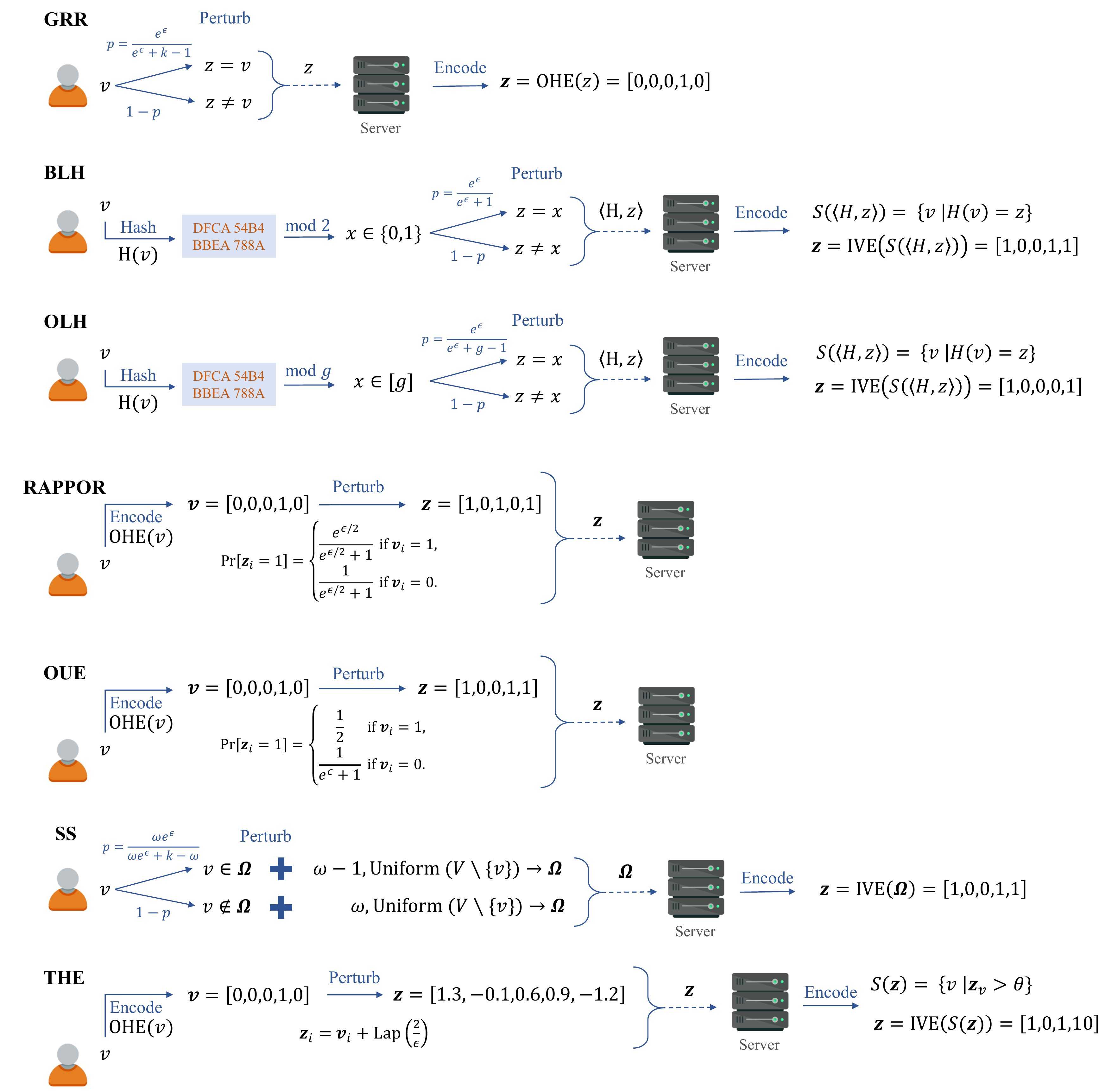}
    \caption{Overview of client-side encoding and perturbation steps for the seven different LDP protocols applied. On the server side, there is also a post-processing step with one-hot encoding (OHE) or indicator vector encoding (IVE), if needed.}
    \label{fig:overview_encoding}
\end{figure}

In addition, each LDP protocol has a different way of encoding and perturbing user's data. 
We thus propose to compare all LDP protocols under the same encoding when training the ML classifier.
More specifically, we will use OHE and Indicator Vector Encoding (IVE)~\cite{ive} as all LDP protocols from Section~\ref{sub:ldp_protocols} are designed for categorical data or discrete data with known domain.
For example, let $\Omega$ be the reported subset of a user after using SS as LDP protocol. Following IVE, we create a binary vector $\mathbf{z}=[b_1,\dots,b_k] \in \{0,1\}^k$ of length $k$, where the $v$-th entry is set to 1 if $v \in \Omega$, and $0$, otherwise. 
In other words, $\mathbf{z}$ represents the subset $\Omega$ in a binary format.
Fig.~\ref{fig:overview_encoding} illustrates the LDP encoding and perturbation at the user side and how to achieve a ``homogeneous encoding" for all the seven LDP protocols at the server side.
Last, all ``non-sensitive" attributes $X$ are encoded using OHE. 

\section{Experimental Evaluation} \label{sec:experimental_evaluation}

In this section, we present our experiments' setting and main results.
Supplementary results can be found in Appendix~\ref{app:add_results}.
Our main Research Questions (RQ) are:
\begin{itemize}
    \item \textbf{RQ1.} Overall, how does preprocessing multidimensional data with $\epsilon$-LDP affect the fairness and utility of ML binary classifiers with the same hyperparameters used before and after sanitization?
    \item \textbf{RQ2.} Which privacy budget-splitting solution leads to less harm to the fairness and utility of an ML binary classifier?
    \item \textbf{RQ3.} How do different LDP protocols affect the fairness and utility of an ML binary classifier, and which one is more suitable for the different real-world scenarios applied?    
\end{itemize}

\subsection{Setup of Experiments} \label{sub:setup_experiments}

\noindent \textbf{General setting.} For all experiments, we consider the following setting:

\begin{itemize}
    \item \textbf{Environment.} All algorithms are implemented in Python 3 with Numpy~\cite{numpy}, Numba~\cite{numba}, and Multi-Freq-LDPy~\cite{multi_freq_ldpy} libraries, and run on a local machine with 2.50GHz Intel Core i9 and 64GB RAM. 
    The codes we develop for all experiments are available in a \textbf{GitHub repository}~\cite{artifact_fairness_ldp}. 

    \item \textbf{ML classifier.} We used the state-of-the-art\footnote{\url{https://www.kaggle.com/kaggle-survey-2022}.} LGBM~\cite{LGBM_paper} as predictor $\hat{Y}$. 

    \item \textbf{Encoding.} We only use discrete and categorical attributes, which are encoded using OHE or IVE (see Fig.~\ref{fig:overview_encoding}) and the target is binary, \ie, $Y \in \{0,1\}$.

    \item \textbf{Training and testing sets.} We randomly select $80\%$ as training set and the remaining $20\%$ as testing set. 
    We apply LDP on the training set only. 
    That is, the samples in the testing set are the original samples (\ie, no LDP).

    \item \textbf{Stability.} Since LDP protocols, train/test splitting, and ML algorithms are randomized, we report average results over 20 runs.
    
\end{itemize}

\noindent \textbf{Datasets.} Table~\ref{tab:datasetsInfo} summarizes all datasets used in our experiments.
For ease of reproducibility, we use real-world and open datasets.

\begin{table}[ht]
\footnotesize
\centering
\caption{Description of the datasets used in the experiments.}
\label{tab:datasetsInfo}
\renewcommand\arraystretch{1.2}
\begin{tabular}{@{}p{2.5cm}@{}@{}p{2cm}@{}@{}p{1.5cm}@{}@{}p{4cm}@{}@{}p{2cm}@{}}\toprule
\emph{Dataset} & \emph{n} \qquad & \emph{$A_{p}$} \qquad & \emph{$A_{s}$, domain size $k$} & \emph{Y} \\\midrule
Adult& $45849$ & gender & - gender, $k=2$ & income\\
&  &   & - race, $k=5$ & \\
&  &   & - native country, $k=41$ & \\
&  &   & - age, $k=74$ & \\
ACSCoverage & $98739$& DIS & - DIS, $k=2$ & PUBCOV\\
&  &   & - AGEP, $k=50$ & \\
&  &   & - SEX, $k=2$ & \\
&  &   & - SCHL, $k=24$ & \\
LSAC & $20427$ & race & - race, $k=2$ & pass bar\\
&  &   & - gender, $k=2$ & \\
&  &   & - family income, $k=5$ & \\
&  &   & - full time, $k=2$ & \\
\bottomrule
\hline
\end{tabular}
\end{table}

\begin{itemize}

    \item \textbf{Adult.} We use $26000$ as threshold to binarize the target variable ``income" of the \emph{reconstructed Adult} dataset~\cite{ding2021retiring}.
    After cleaning, $n=45849$ samples are kept.
    We excluded ``capital-gain" and ``capital-loss" and used the remaining $10$ discrete and categorical attributes.
    We considered $A_{s}=\{\textrm{gender, race, native-country, age}\}$ as sensitive attributes for LDP sanitization and $A_{p}=\textrm{gender}$ as the protected attribute for fairness assessment.

    \item \textbf{ACSCoverage.} This dataset\footnote{The full documentation for the description of all attributes is in \url{https://www.census.gov/programs-surveys/acs/microdata/documentation.html}.} is retrieved with the \texttt{folktables}~\cite{ding2021retiring} Python package and the binary target ``PUBCOV" designates whether an individual is covered by public health insurance or not. 
    We select the year 2018 and the ``Texas'' state, with $n=98739$ samples. 
    We removed ``DEAR", ``DEYE", ``DREM", and ``PINCP" and used the remaining $15$ discrete and categorical attributes.
    We considered $A_{s}=\{\textrm{DIS, AGEP, SEX, SCHL}\}$ as sensitive attributes for LDP sanitization and $A_{p}=\textrm{DIS}$ as the protected attribute (\ie, disability) for fairness assessment. 
    
    \item \textbf{LSAC.} This dataset is from the Law School Admissions Council (LSAC) National Bar Passage Study~\cite{wightman1998lsac} and the binary target ``pass\_bar" indicates whether or not a candidate has passed the bar exam.
    After cleaning, $n=20427$ samples are kept.
    We only consider as attributes: `gender', `race', `family income', `full time', `undergrad GPA score' (discretized to $\{1.5, 2.0, ..., 4.5\}$), and `LSAT score' (rounded to the closest integer). 
    The `race' attribute was binarized to \{black, other\}.
    We set $A_{s}=\{\textrm{race, gender, family income, full time}\}$ as sensitive attributes for LDP sanitization and $A_{p}=\textrm{race}$ as the protected attribute for fairness assessment. 

\end{itemize}

\noindent \textbf{Evaluated methods.} The methods we use and compare are:

\begin{itemize}
    \item \textbf{(Baseline) NonDP.} This is our baseline with LGBM trained over original data (\ie, $D=(X,A_{s},Y)$).
    We searched for the best hyperparameters using Bayesian optimization~\cite{hyperopt2013} through $100$ iterations varying: $max\_depth \in [3, 50]$, $n\_estimators \in [50, 2000]$, and $learning\_rate \in (0.01, 0.25)$;

    \item \textbf{LDP protocols.} We pre-processed $Z_{s}=\calM(A_{s})$ of the training sets using all seven LDP protocols from Section~\ref{sub:ldp_protocols} (\ie, GRR, RAPPOR, OUE, SS, BLH, OLH, and THE) as $\calM$.
    We used the best hyperparameters found for the NonDP model and trained LGBM over $D_{z}=(X,Z_{s},Y)$.
    For all datasets, we set $d_s$ to $4$. That is, $d_s = |A_{s}|=4$. 
    To satisfy $\epsilon$-LDP (\cf{} Definition~\ref{prop:seq_comp}), we split the privacy budget $\epsilon$ following the two solutions described in Section~\ref{sec:problem_setting_methodology} (\ie, the state-of-the-art uniform and our k-based solution). 

\end{itemize}

\noindent \textbf{Metrics.} We evaluate the performance of LGBM trained over the original data (\ie, NonDP baseline) and LDP-based data on privacy, utility, and fairness: 

\begin{itemize}

    \item \textbf{Privacy.} We vary the privacy parameter in the range of $\epsilon = \{0.25, 0.5, 1, 2, 4, 8, 10, 20, 50\}$.
    At $\epsilon = 0.25$ the ratio of probabilities is bounded by $e^{0.25} \approx 1.3$ giving nearly indistinguishable distributions, whereas at $\epsilon = 50$ almost no privacy is guaranteed. 
    
    \item \textbf{Utility.} We use accuracy (acc), f1-score (f1), area under the receiver operating characteristic curve (auc), and recall as utility metrics;

    \item \textbf{Fairness.} We use the metrics of Section~\ref{sub:fairness_metrics} (\ie, DI, SPD, EOD, and OAD).

\end{itemize}

\subsection{Main Results} \label{sub:results}

\begin{figure}
    \centering
    \includegraphics[width=1\linewidth]{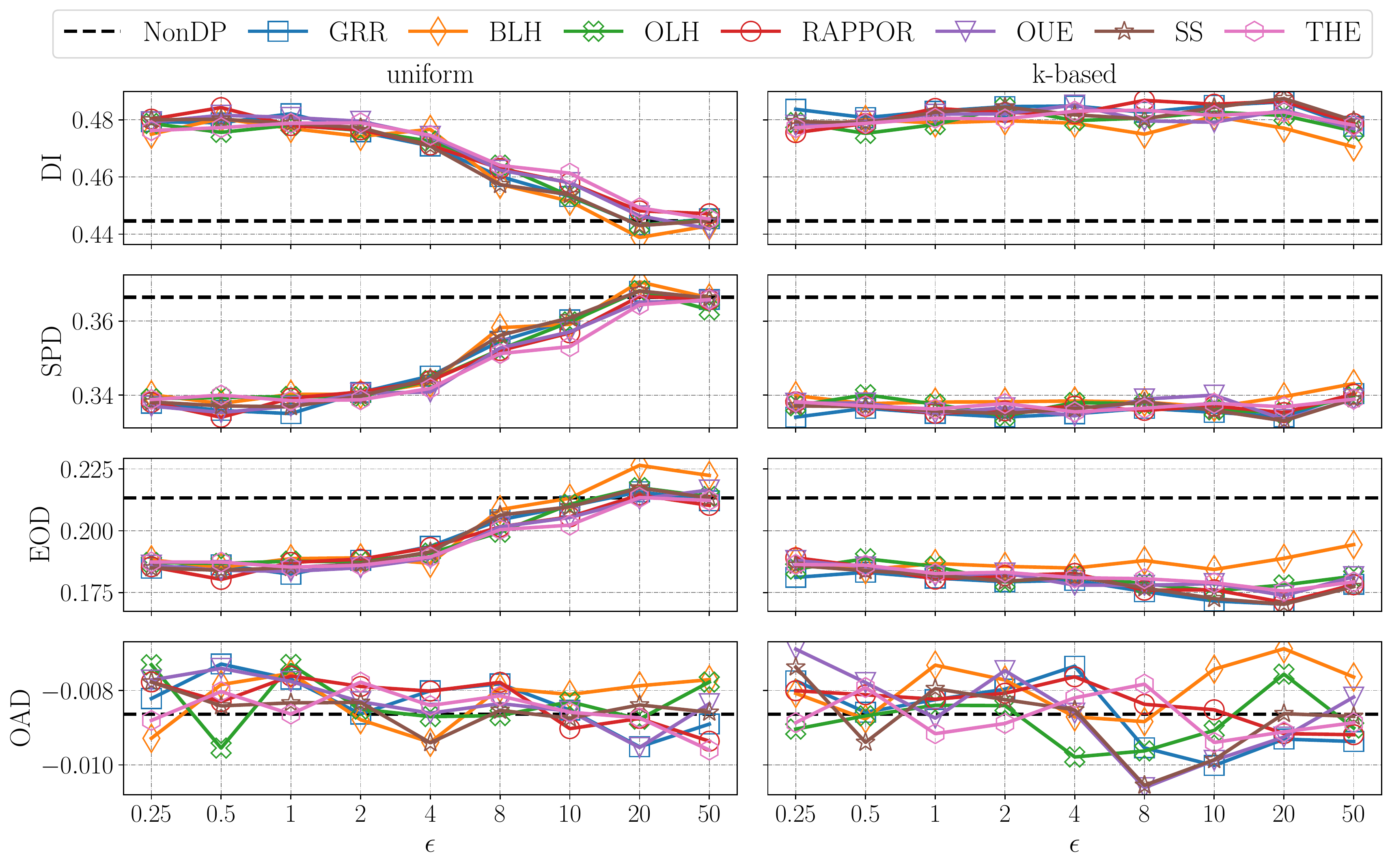}
    \caption{Fairness metrics (y-axis) by varying the privacy guarantees (x-axis), the $\epsilon$-LDP protocol, and the privacy budget splitting solution (\ie, uniform on the left-side and our k-based on the right-side), on the Adult~\cite{ding2021retiring} dataset.}
    \label{fig:fairness_adult}
\end{figure}

\begin{figure}
    \centering
    \includegraphics[width=1\linewidth]{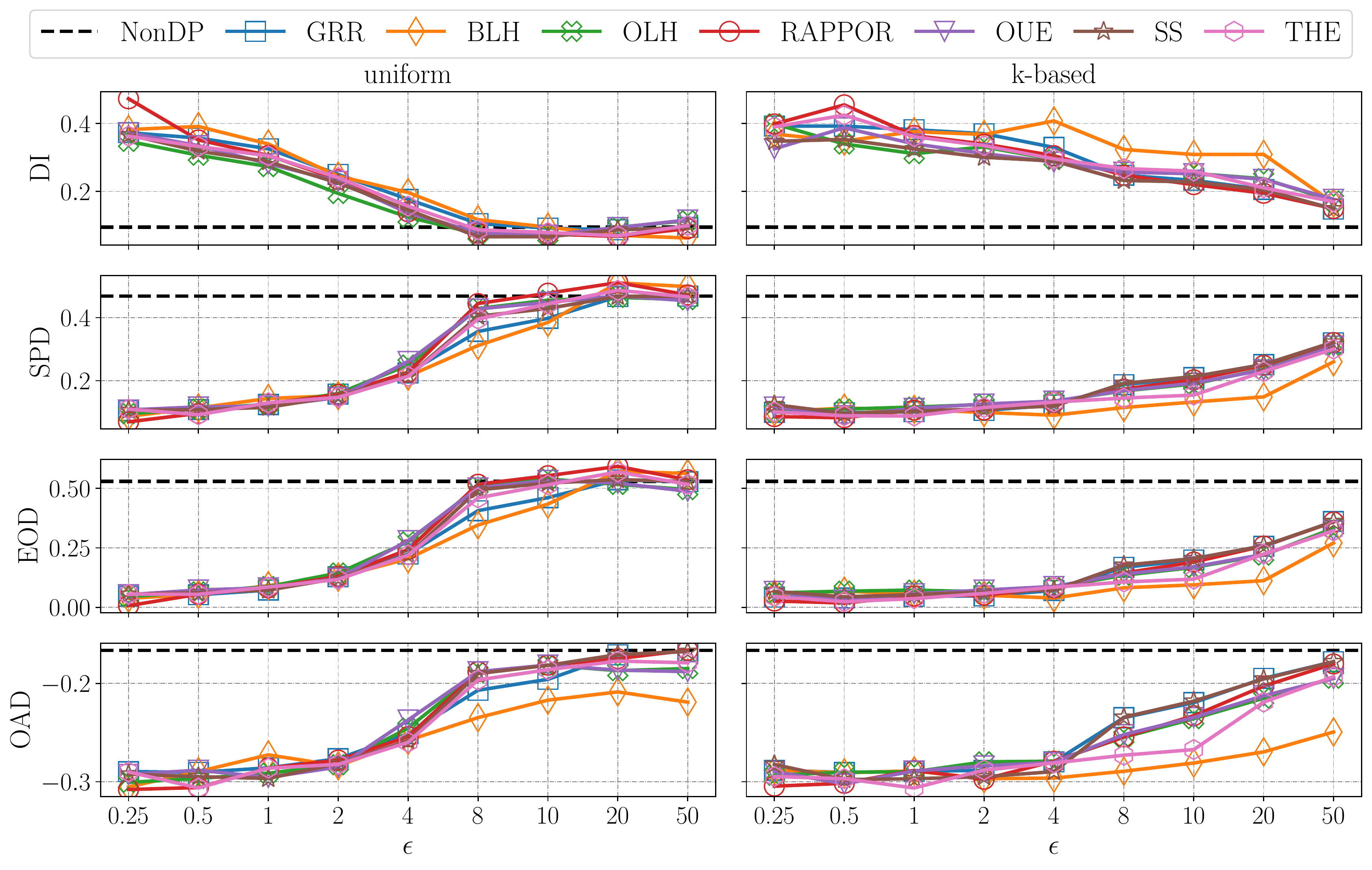}
    \caption{Fairness metrics (y-axis) by varying the privacy guarantees (x-axis), the $\epsilon$-LDP protocol, and the privacy budget splitting solution (\ie, uniform on the left-side and our k-based on the right-side), on the ACSCoverage~\cite{ding2021retiring} dataset.}
    \label{fig:fairness_acs}
\end{figure}

\begin{figure}
    \centering
    \includegraphics[width=1\linewidth]{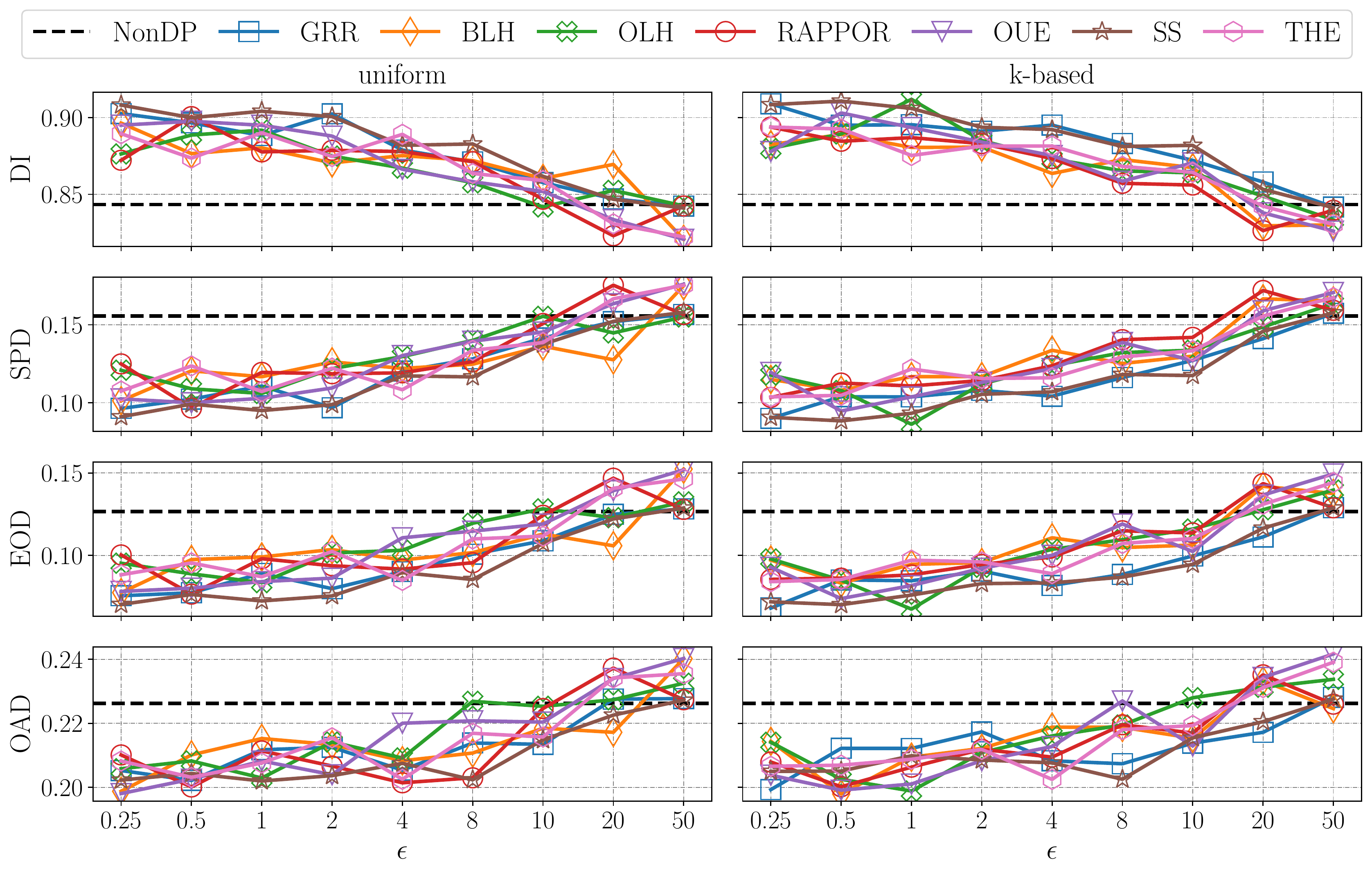}
    \caption{Fairness metrics (y-axis) by varying the privacy guarantees (x-axis), the $\epsilon$-LDP protocol, and the privacy budget splitting solution (\ie, uniform on the left-side and our k-based on the right-side), on the LSAC~\cite{wightman1998lsac} dataset.}
    \label{fig:fairness_lsac}
\end{figure}

\noindent \textbf{LDP impact on fairness.} Fig.~\ref{fig:fairness_adult} (Adult), Fig.~\ref{fig:fairness_acs} (ACSCoverage), and Fig.~\ref{fig:fairness_lsac} (LSAC) illustrate the privacy-fairness trade-off for the NonDP baseline and all the seven LDP protocols, considering both uniform and our k-based privacy budget splitting solutions.
From these figures, one can notice that fairness is, in general, slightly improved for all seven LDP protocols under both the uniform and the k-based solution. 
For instance, for the DI metric in Fig.~\ref{fig:fairness_adult}, the Non-DP data indicates a value of 0.44 showing discrimination against women and, by applying LDP protocols, DI tended to increase to $\sim$0.48  (with $\epsilon$ = 0.25) resulting in a slight improvement in fairness. Similarly, SPD decreased from 0.37 to $\sim$0.34 after applying LDP protocols. The same behavior is obtained for EOD.
The exception was in Fig.~\ref{fig:fairness_acs} for the OAD metric in which the gap between privileged and unprivileged groups was accentuated (favoring the unprivileged group). 
More specifically, the NonDP baseline has OAD equal to -0.17, and after satisfying LDP for both uniform and k-based solutions and using all LDP protocols, the gap between the privileged and unprivileged groups increased to -0.3. 
In other words, we start with favoritism towards the unprivileged group (negative value) and this favoritism increased after LDP.

Note also that when applying the uniform privacy budget splitting solution (see left-side plots), all fairness metrics were less robust to LDP than our k-based solution and, thus, returned to the NonDP baseline value in low privacy regimes. 
With our k-based solution (see right-side plots), all fairness metrics continued to be slightly better for all privacy regimes for the Adult dataset in Fig.~\ref{fig:fairness_adult}.
For the ACSCoverage dataset, not all fairness metrics returned to the NonDP baseline value and for the LSAC dataset, a similar behavior was noticed for both uniform and k-based solutions.
These differences are mainly influenced by the domain size $k$ of the sensitive attributes.
For instance, while Adult has sensitive attributes with higher values of $k$, LSAC has many binary sensitive attributes.

\begin{figure}
    \centering
    \includegraphics[width=1\linewidth]{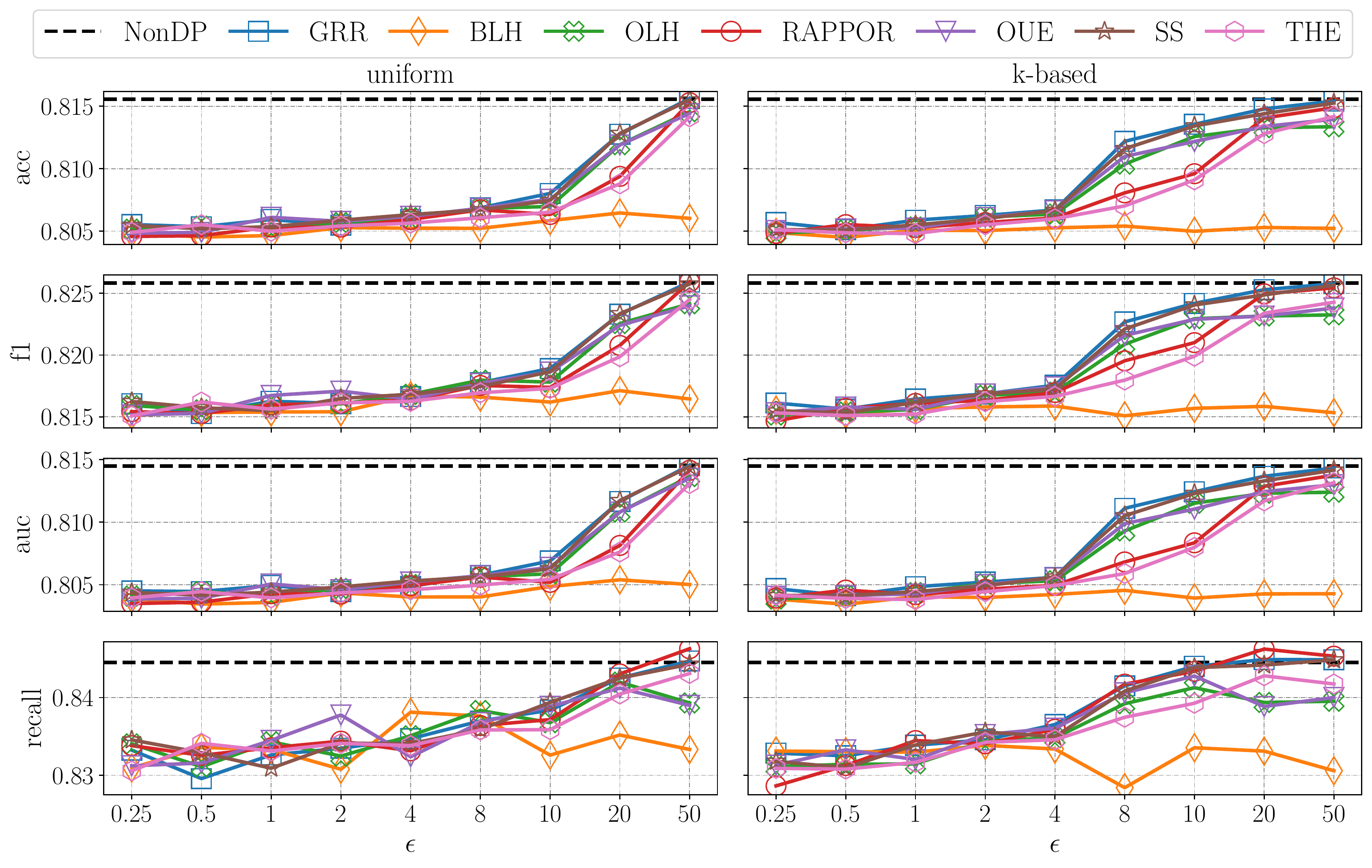}
    \caption{Utility metrics (y-axis) by varying the privacy guarantees (x-axis), the $\epsilon$-LDP protocol, and the privacy budget splitting solution (\ie, uniform on the left-side and our k-based on the right-side), on the Adult~\cite{ding2021retiring} dataset. }
    \label{fig:utility_adult}
\end{figure}

\begin{figure}
    \centering
    \includegraphics[width=1\linewidth]{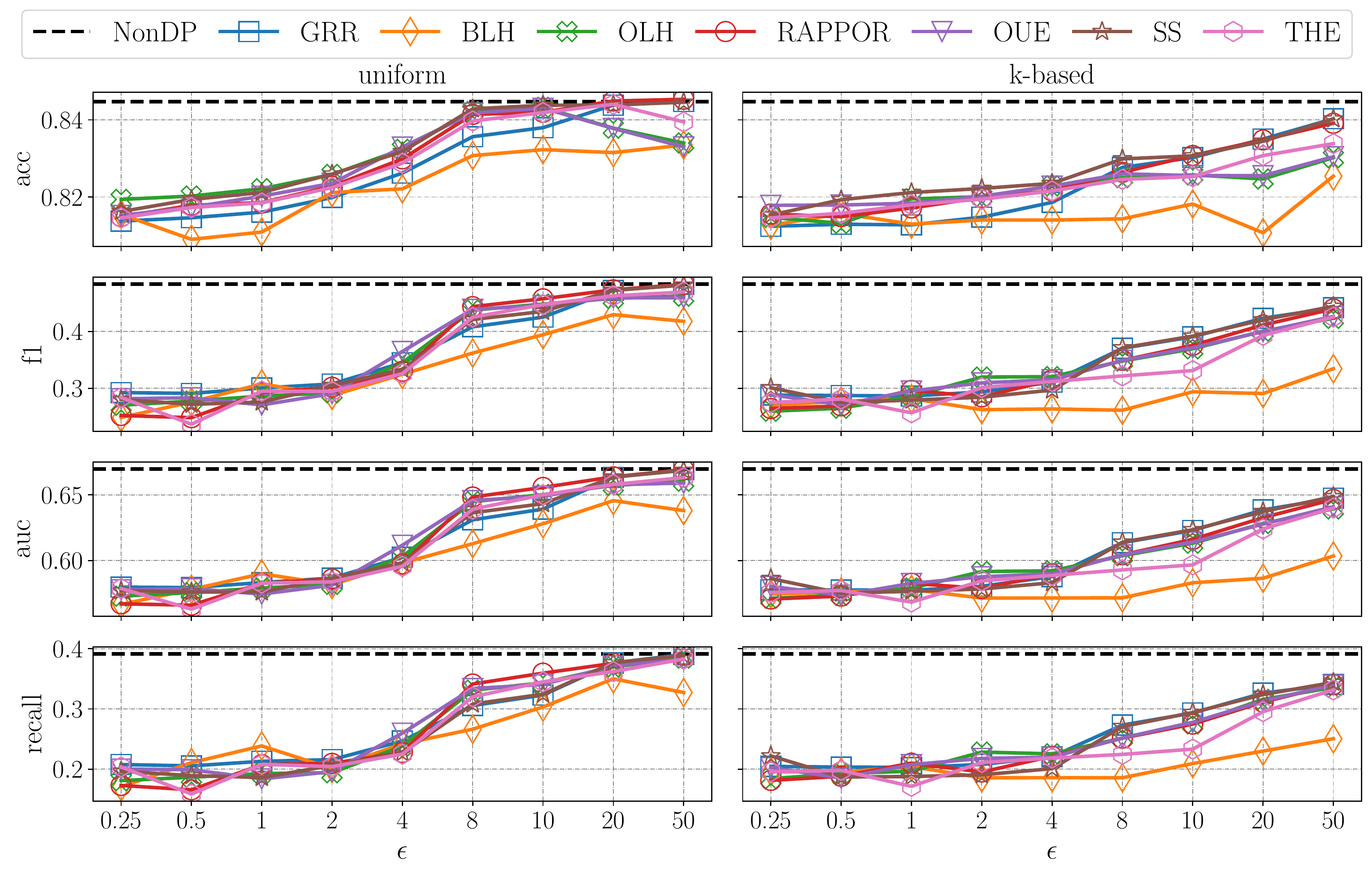}
    \caption{Utility metrics (y-axis) by varying the privacy guarantees (x-axis), the $\epsilon$-LDP protocol, and the privacy budget splitting solution (\ie, uniform on the left-side and our k-based on the right-side), on the ACSCoverage~\cite{ding2021retiring} dataset.}
    \label{fig:utility_acs}
\end{figure}

\begin{figure}
    \centering
    \includegraphics[width=1\linewidth]{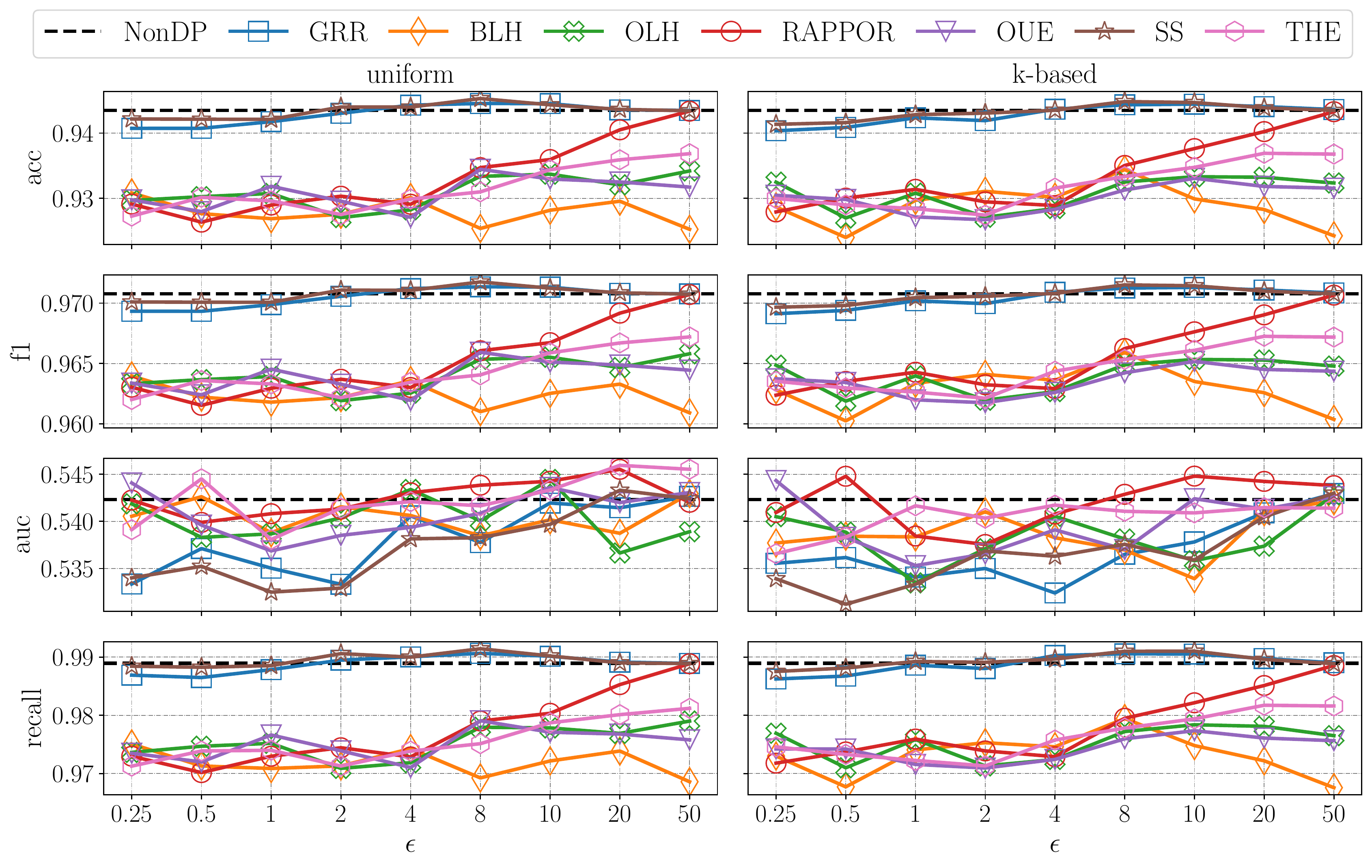}
    \caption{Utility metrics (y-axis) by varying the privacy guarantees (x-axis), the $\epsilon$-LDP protocol, and the privacy budget splitting solution (\ie, uniform on the left-side and our k-based on the right-side), on the LSAC~\cite{wightman1998lsac} dataset.}
    \label{fig:utility_lsac}
\end{figure}

\noindent \textbf{LDP impact on utility.} Fig.~\ref{fig:utility_adult} (Adult), Fig.~\ref{fig:utility_acs} (ACSCoverage), and Fig.~\ref{fig:utility_lsac} (LSAC) illustrate the privacy-utility trade-off for the NonDP baseline and all the seven LDP protocols, considering both uniform and our k-based privacy budget splitting solutions. 
From these figures, one can note that, in general, the impact of $\epsilon$-LDP on utility metrics is minor. For instance, for the Adult dataset (Fig.~\ref{fig:utility_adult}), only $\sim 1\%$ of utility loss for all metrics is observed.
Regarding privacy budget splitting, for the Adult dataset, our k-based solution is more robust to LDP as it only drops in higher privacy regimes (\ie, smaller $\epsilon$ values) than the uniform solution.
One main explanation for this behavior is because there is more discrepancy in the domain size $k$'s of the sensitive attributes $A_{s}$ and, consequently, more privacy budget $\epsilon$ are allocated to those attributes with high $k$.
For this reason, the uniform solution preserved more utility for the ACSCoverage dataset in Fig.~\ref{fig:utility_acs}, and both solutions had similar results for the LSAC dataset in Fig.~\ref{fig:utility_lsac} due to sensitive attributes with small domain size $k$.

\noindent \textbf{Summary.} We summarize our main findings for the three research questions formulated at the beginning of Section~\ref{sec:experimental_evaluation}. 
We highlight these findings are generic and were also confirmed in additional experiments presented in Appendix~\ref{app:add_results}.
\textbf{(RQ1)} Using the same hypeparameters configuration, $\epsilon$-LDP positively affects fairness in ML (see Figs.~\ref{fig:fairness_adult}--\ref{fig:fairness_lsac}) while having a negligible impact on model's utility (see Figs.~\ref{fig:utility_adult}--\ref{fig:utility_lsac}).
This contrasts the findings of~\cite{Bagdasaryan2019,Ganev2022} that state that under the same hyperparameters configuration, $\epsilon$-DP negatively impacts fairness. 
Although the aforementioned research works concern \textit{gradient perturbation} in central DP, the recent work of de Oliveira \etal~\cite{deOliveira2023} has shown that when searching for the best hyperparameters for both non-private and DP models, the $\epsilon$-DP impact on fairness is negligible.
In our case, we focused on \textit{input perturbation}, \ie, randomizing multiple sensitive attributes before training any ML algorithm, and discovered a positive impact of $\epsilon$-(L)DP on fairness.
\textbf{(RQ2)} Our k-based solution consistently led to better fairness than the state-of-the-art uniform solution when there exist sensitive attributes with high domain size $k$ (\eg, for both Adult and ACSCoverage datasets).
Naturally, when all sensitive attributes have a binary domain, our k-based solution is equivalent to the uniform solution.
For this reason, both state-of-the-art uniform and our k-based solution led to similar privacy-utility-fairness trade-off for the LSAC dataset (see Figs.~\ref{fig:fairness_lsac} and~\ref{fig:utility_lsac}).
Therefore, regarding utility, k-based is better when sensitive attributes have higher domain sizes $k$, which coincides with real-world data collections.
\textbf{(RQ3)} In general, GRR and SS presented the best privacy-utility-fairness trade-off for all three datasets. 
This is because GRR has only one perturbed output value and because SS is equivalent to GRR when $\omega=1$, thus, not introducing inconsistencies for a user's profile.
The term \textit{inconsistency} refers to an user being multiple categories in a given attribute, \ie, being both woman and man at the same time.
In fact, this is precisely what happens with UE protocols that perturb each bit independently or with LH protocols in which many values can hash to the same perturbed value.
For this reason, since BLH hashes the input set $V \to \{0,1\}$, it consistently presented the worst utility results for all three datasets, and only for ACSCoverage (see Fig.~\ref{fig:fairness_acs}), it presented slightly better fairness results than all other LDP protocols.

\section{Conclusion and Perspectives}\label{sec:conclusion}

This paper presented an in-depth empirical study of the impact of pre-processing multidimensional data with seven state-of-the-art $\epsilon$-LDP protocols on fairness and utility in binary classification tasks.
In our experiments, GRR~\cite{kairouz2016discrete} and SS~\cite{wang2016mutual,Min2018} presented the best privacy-utility-fairness trade-off than RAPPOR~\cite{rappor}, OUE~\cite{tianhao2017}, THE~\cite{tianhao2017}, BLH~\cite{Bassily2015}, and OLH~\cite{tianhao2017}.
In addition, we proposed a new privacy budget splitting solution named k-based, which generally led to better fairness and performance results than the state-of-the-art solution that splits $\epsilon$ uniformly~\cite{Arcolezi2021_rs_fd,wang2019}.
Globally, while previous research~\cite{Bagdasaryan2019,Ganev2022} has highlighted that DP worsens fairness in ML under the same hyperparameter configuration, our study finds that LDP slightly improves fairness and does not significantly impair utility.
Indeed, there is still much to explore in the area of privacy-fairness-aware ML, and this study's empirical results can serve as a basis for future research directions.
For instance, we intend to formally investigate the privacy-utility-fairness trade-off on binary classification tasks when varying the distribution of the protected attribute, the target, and their joint, and propose new methods accordingly.
Last, we plan to investigate the impact of LDP pre-processing on different ML algorithms, such as deep neural networks.

\subsubsection*{Acknowledgements} 
This work was supported by the European Research Council (ERC) project HYPATIA under the European Union’s Horizon 2020 research and innovation programme. Grant agreement n. 835294.

%
%
%

\bibliographystyle{splncs04}
\bibliography{ms.bib}

\appendix

\section{Additional Experiments} \label{app:add_results}

To validate our findings that LDP can improve fairness without sacrificing much utility, we conducted an additional series of experiments by considering a dynamic number of sensitive attributes $d_s$. 
Specifically, for each iteration of the 20 runs (for stability), using a specific dataset such as Adult~\cite{ding2021retiring}, ACSCoverage~\cite{ding2021retiring}, or LSAC~\cite{wightman1998lsac}, we randomly determined the number of sensitive attributes $2\leq d_s \leq 6$, ensuring that the protected attribute $A_p$ is always included in $A_s$, \ie, $A_p \in A_s$.

\begin{figure}
    \centering
    \includegraphics[width=0.95\linewidth]{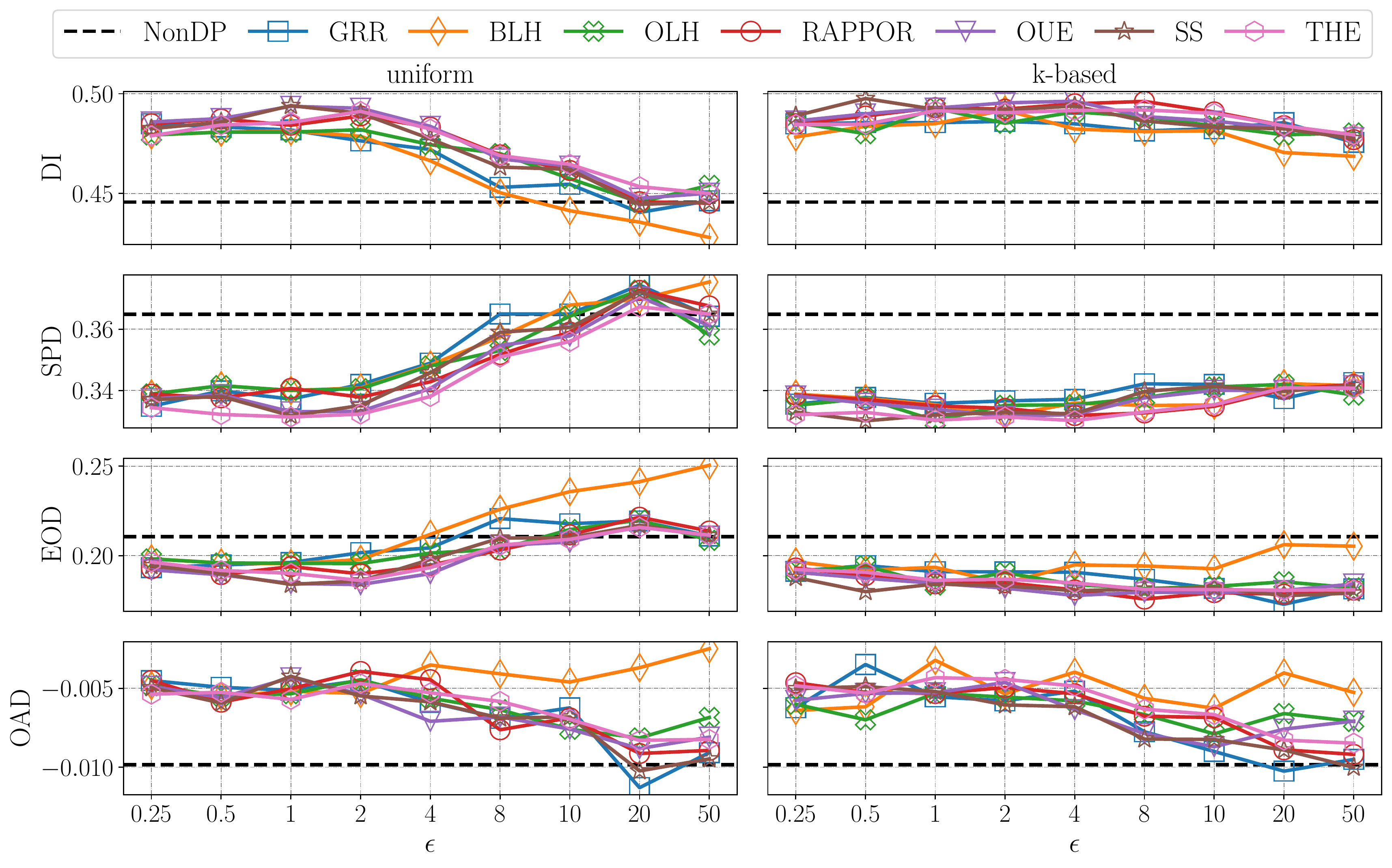}
    \caption{Fairness metrics (y-axis) by varying the privacy guarantees (x-axis), the $\epsilon$-LDP protocol, and the privacy budget splitting solution (\ie, uniform on the left-side and our k-based on the right-side), on the Adult~\cite{ding2021retiring} dataset. The number of sensitive attributes $2\leq d_s \leq 6$ is selected uniformly at random.}
    \label{fig:fairness_adult_appendix}
\end{figure}

\begin{figure}
    \centering
    \includegraphics[width=0.95\linewidth]{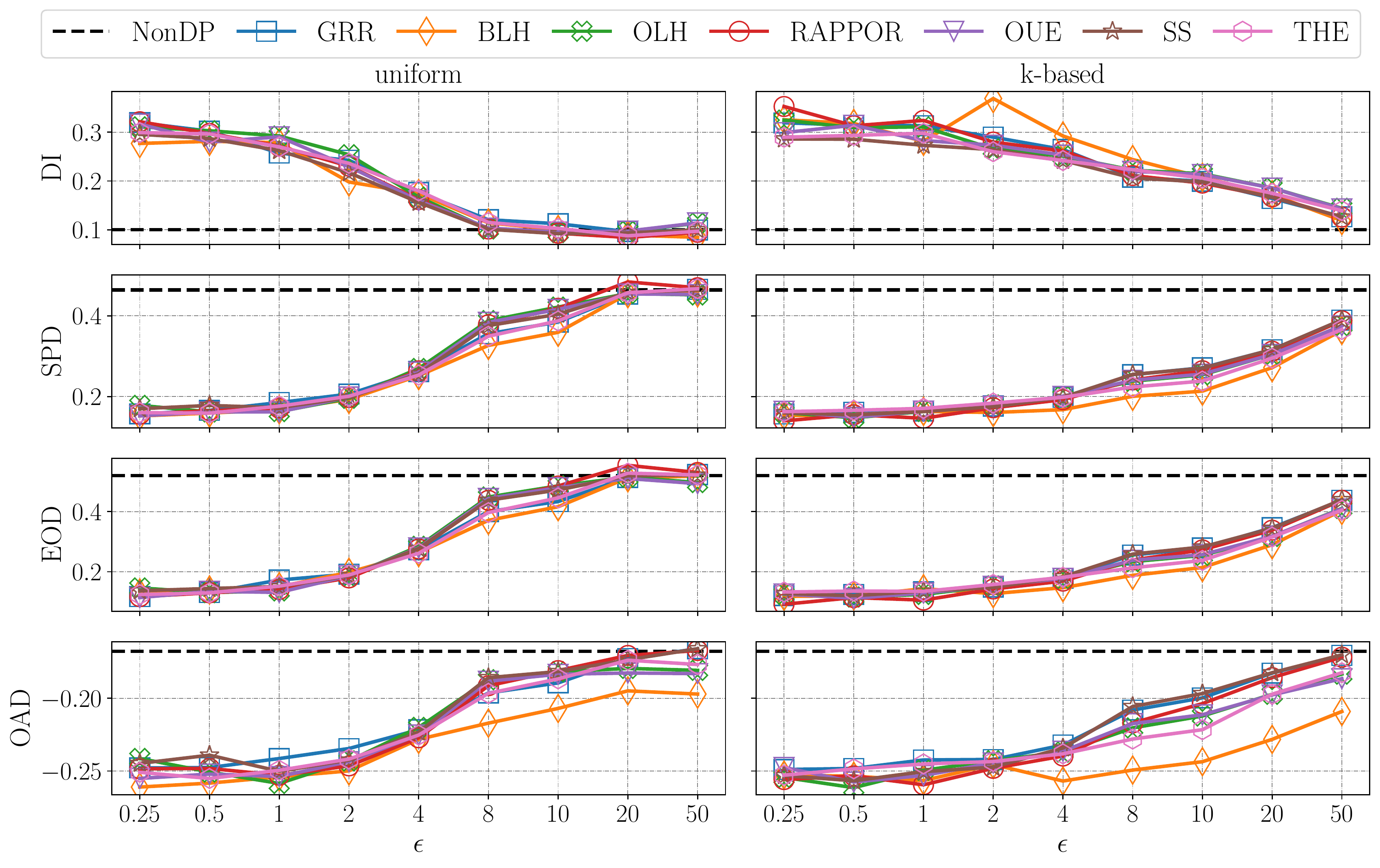}
    \caption{Fairness metrics (y-axis) by varying the privacy guarantees (x-axis), the $\epsilon$-LDP protocol, and the privacy budget splitting solution (\ie, uniform on the left-side and our k-based on the right-side), on the ACSCoverage~\cite{ding2021retiring} dataset. The number of sensitive attributes $2\leq d_s \leq 6$ is selected uniformly at random.}
    \label{fig:fairness_acs_appendix}
\end{figure}

\begin{figure}
    \centering
    \includegraphics[width=0.95\linewidth]{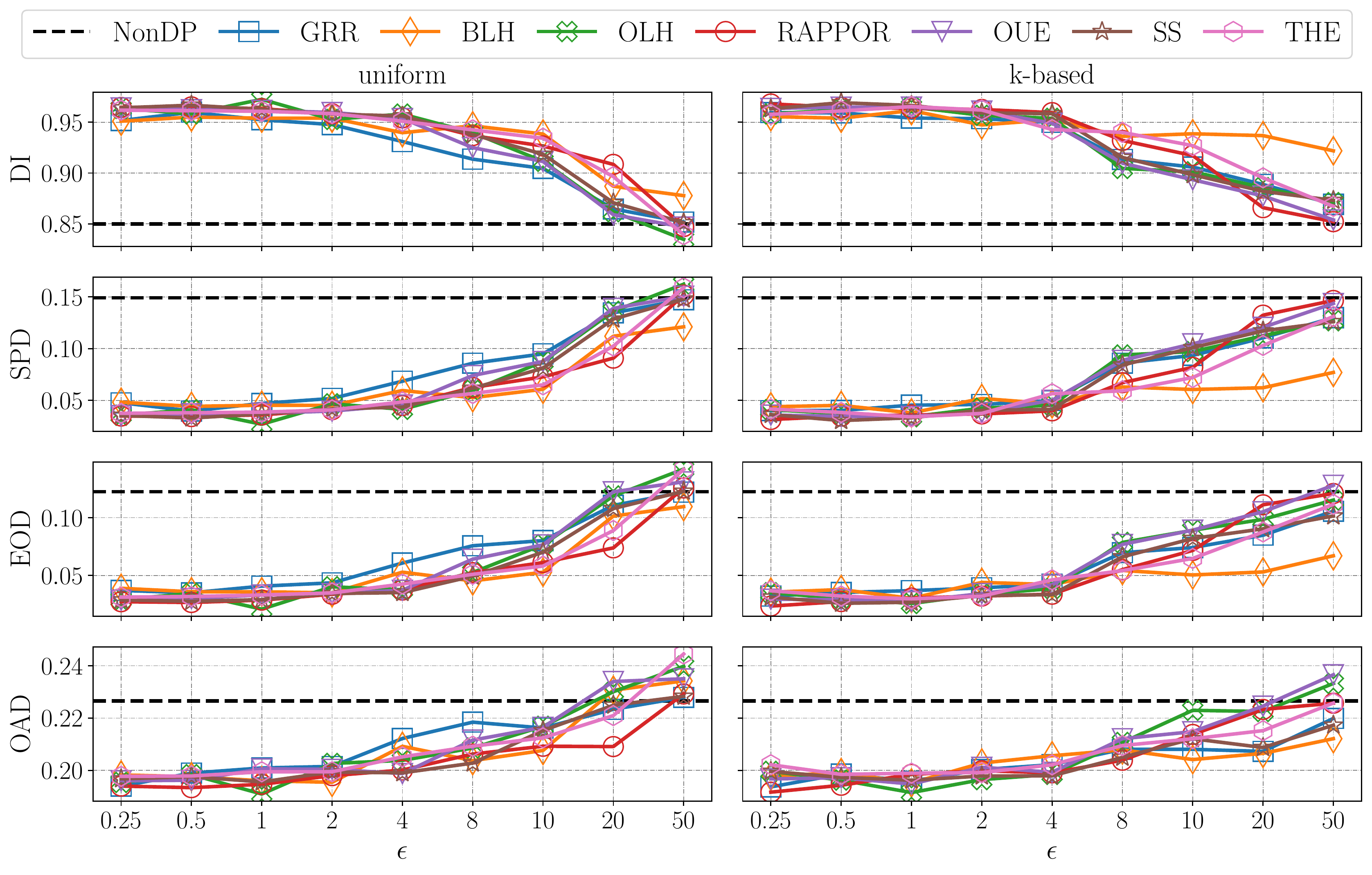}
    \caption{Fairness metrics (y-axis) by varying the privacy guarantees (x-axis), the $\epsilon$-LDP protocol, and the privacy budget splitting solution (\ie, uniform on the left-side and our k-based on the right-side), on the LSAC~\cite{wightman1998lsac} dataset. The number of sensitive attributes $2\leq d_s \leq 6$ is selected uniformly at random.}
    \label{fig:fairness_lsac_appendix}
\end{figure}

\begin{figure}
    \centering
    \includegraphics[width=0.95\linewidth]{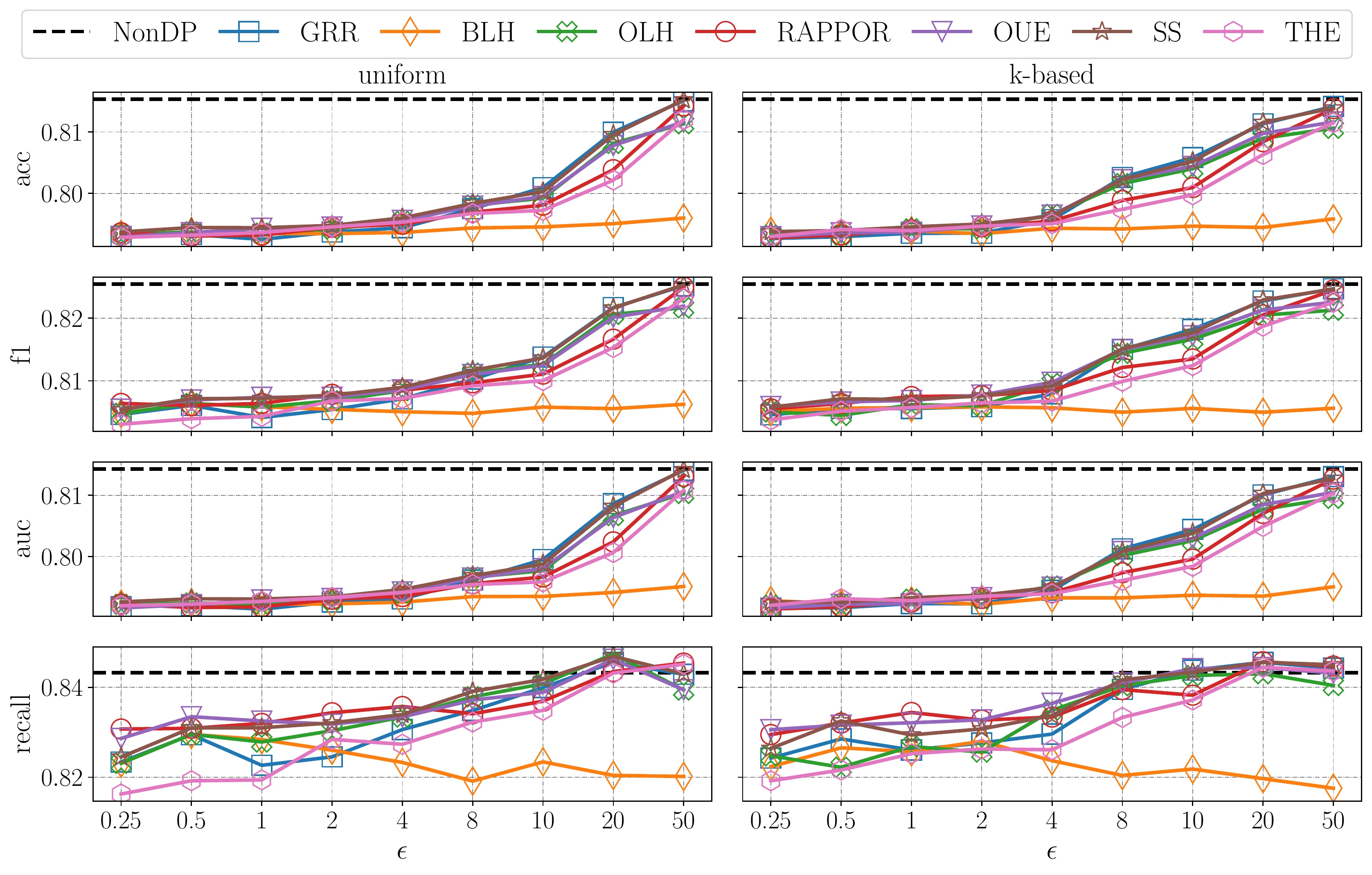}
    \caption{Utility metrics (y-axis) by varying the privacy guarantees (x-axis), the $\epsilon$-LDP protocol, and the privacy budget splitting solution (\ie, uniform on the left-side and our k-based on the right-side), on the Adult~\cite{ding2021retiring} dataset. The number of sensitive attributes $2\leq d_s \leq 6$ is selected uniformly at random.}
    \label{fig:utility_adult_appendix}
\end{figure}

\begin{figure}
    \centering
    \includegraphics[width=0.95\linewidth]{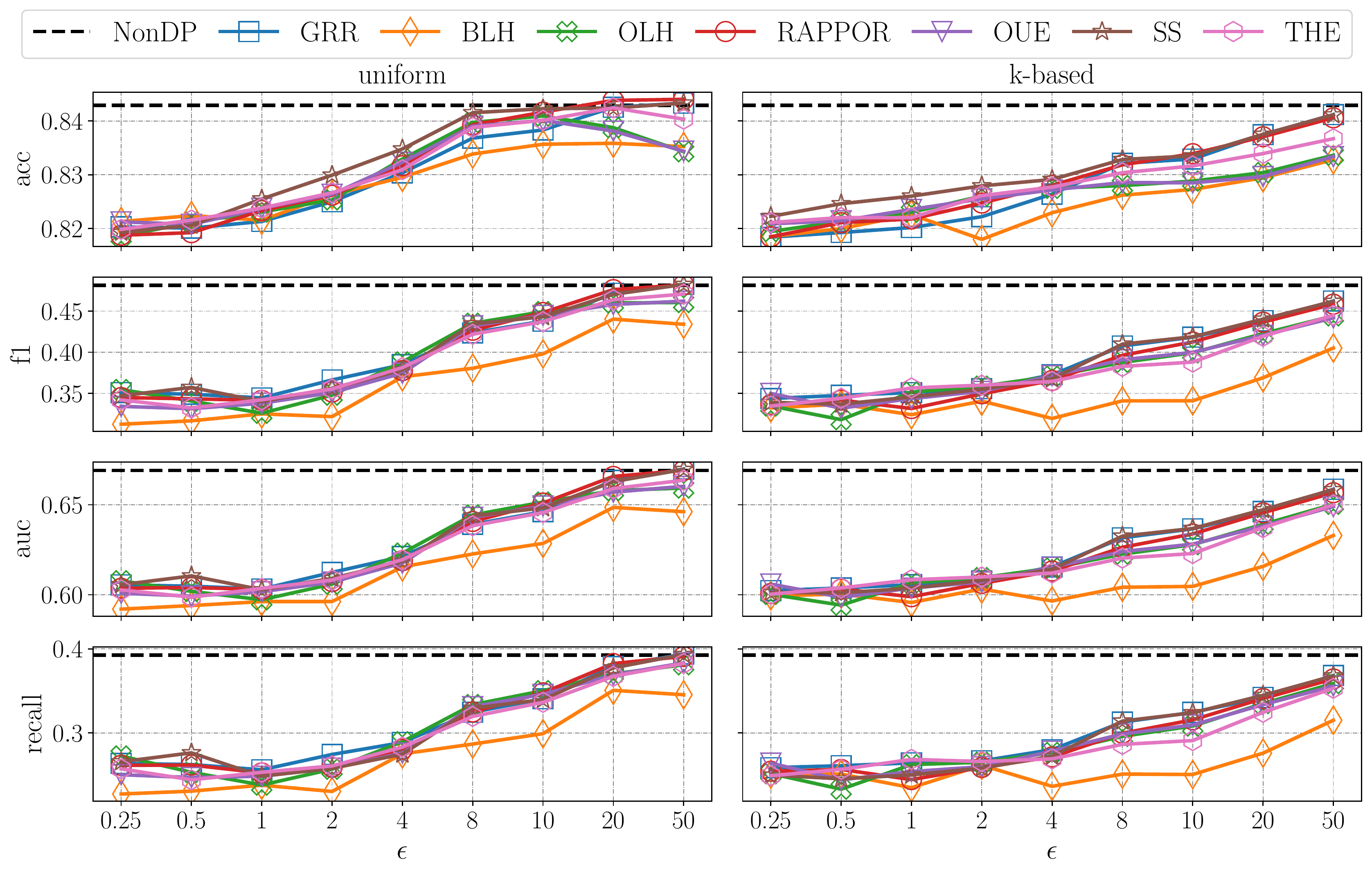}
    \caption{Utility metrics (y-axis) by varying the privacy guarantees (x-axis), the $\epsilon$-LDP protocol, and the privacy budget splitting solution (\ie, uniform on the left-side and our k-based on the right-side), on the ACSCoverage~\cite{ding2021retiring} dataset. The number of sensitive attributes $2\leq d_s \leq 6$ is selected uniformly at random.}
    \label{fig:utility_acs_appendix}
\end{figure}

\begin{figure}
    \centering
    \includegraphics[width=0.95\linewidth]{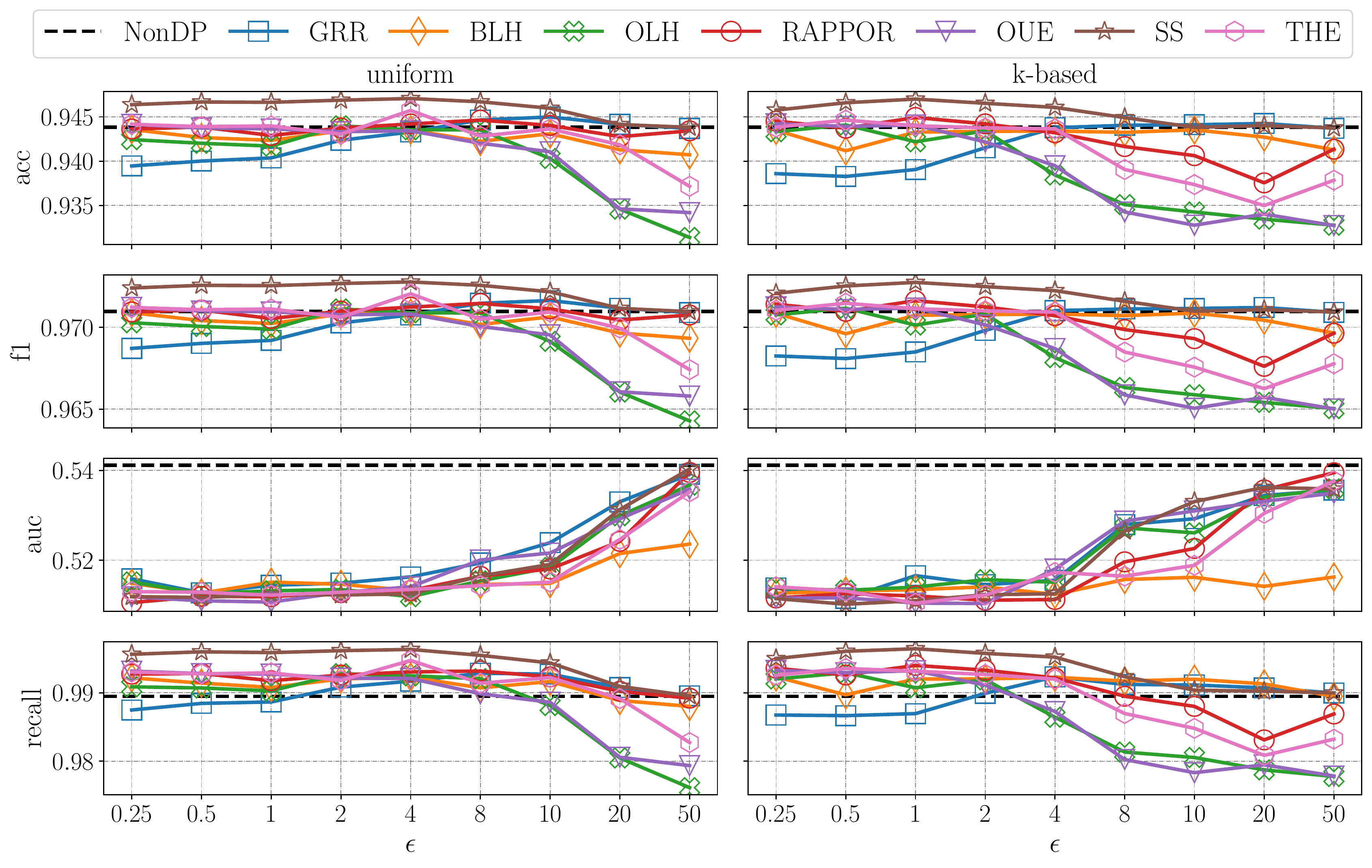}
    \caption{Utility metrics (y-axis) by varying the privacy guarantees (x-axis), the $\epsilon$-LDP protocol, and the privacy budget splitting solution (\ie, uniform on the left-side and our k-based on the right-side), on the LSAC~\cite{wightman1998lsac} dataset. The number of sensitive attributes $2\leq d_s \leq 6$ is selected uniformly at random.}
    \label{fig:utility_lsac_appendix}
\end{figure}

Similar to Figs.~\ref{fig:fairness_adult}--\ref{fig:fairness_lsac} (LDP impact on fairness) and Figs.~\ref{fig:utility_adult}--\ref{fig:utility_lsac} (LDP impact on utility), Figs.~\ref{fig:fairness_adult_appendix}--\ref{fig:utility_lsac_appendix} illustrate the privacy-fairness-utility trade-offs for the Adult, ACSCoverage, and LSAC dataset, respectively. 
These figures consider the NonDP baseline and the seven LDP protocols, as well as both the uniform and our k-based privacy budget splitting solutions.
From Figs.~\ref{fig:fairness_adult_appendix}--\ref{fig:utility_lsac_appendix}, one can observe that the results follow similar trends as those presented in Section~\ref{sec:experimental_evaluation}.
Specifically, the LDP pre-processing positively affects fairness while only having a minor impact on the utility of the ML model.

\end{document}